\pdfoutput=1
\documentclass{article}

    \PassOptionsToPackage{numbers, compress}{natbib}

    \usepackage[preprint]{neurips_data_2024}

\usepackage[utf8]{inputenc} 
\usepackage[T1]{fontenc}    
\usepackage{hyperref}       
\usepackage{url}            
\usepackage{booktabs}       
\usepackage{amsfonts}       
\usepackage{nicefrac}       
\usepackage{microtype}      
\usepackage{xcolor}         
\usepackage{amsmath}        
\usepackage{enumitem}       
\usepackage{multirow}       
\usepackage[pdftex]{graphicx}
\usepackage{bbm}
\title{Inpainting the Gaps: A Novel Framework for Evaluating Explanation Methods in Vision Transformers}

\author{%
Lokesh Badisa$^1$ \quad
Sumohana S. Channappayya$^1$\\
$^1$Indian Institute of Technology Hyderabad\\
\texttt{ai21btech11005@iith.ac.in,sumohana@ee.iith.ac.in}
}

\begin{document}

\maketitle

\begin{abstract}
The perturbation test remains the go-to evaluation approach for explanation methods in computer vision. This evaluation method has a major drawback of test-time distribution shift due to pixel-masking that is not present in the training set. To overcome this drawback, we propose a novel evaluation framework called \textbf{Inpainting the Gaps (InG)}. Specifically, we propose inpainting parts that constitute partial or complete objects in an image. In this way, one can perform meaningful image perturbations with lower test-time distribution shifts, thereby improving the efficacy of the perturbation test. InG is applied to the PartImageNet dataset to evaluate the performance of popular explanation methods for three training strategies of the Vision Transformer (ViT). Based on this evaluation, we found Beyond Intuition and Generic Attribution to be the two most consistent explanation models. Further, and interestingly, the proposed framework results in higher and more consistent evaluation scores across all the ViT models considered in this work. To the best of our knowledge, InG is the first semi-synthetic framework for the evaluation of ViT explanation methods.
\end{abstract}

 \section{Introduction}
Transformer models dominate the field of natural language processing (NLP) \cite{vaswani2017attention}. With their success in NLP, transformer models have been adapted for computer vision tasks and are called Vision Transformers (ViT)~\cite{dosovitskiy2020image}. ViTs deliver state-of-the-art performance on various computer vision tasks and are preferred to convolutional neural networks (CNNs) in many vision applications~\cite{radford2021learning, peebles2023scalable}. While ViTs have quickly become popular, methods for explaining their prediction are still nascent. This lack of explainability hinders their adoption in critical applications such as healthcare and criminal law that require explainability. CNNs, on the other hand, have benefited from a long line of research on their explainability. 

There has been a spurt in explainability methods for ViTs \cite{kashefi2023explainability}. To choose between the multitude of explanations, notions like completeness, correctness, etc. \cite{nauta2023anecdotal} have been proposed in the literature. These aspects evaluate the explanation method at a pixel level (disconnected from the downstream task \cite{hesse2023funnybirds}). In pixel-level methods,  pixel-masking is the dominant paradigm where a pixel's importance is determined by masking it out. However, pixel masking introduces a distribution shift at test time, thereby bringing into question the fairness of this approach. To overcome this problem, FunnyBirds\cite{hesse2023funnybirds} proposed a synthetic dataset where part perturbations are available in the training set. This method trains models on this synthetic dataset where test samples being out-of-distribution (OOD) is not possible as they are present in the training set. However, the FunnyBirds method requires training the model on their synthetic dataset. It is also not obvious how this method translates to real-world images. 

Inspired by the FunnyBirds work, we create a semi-synthetic dataset of real-world images where part perturbations are incorporated by inpainting the parts. We use a recently proposed part-level dataset, PartImageNet\cite{he2022partimagenet}, where part annotations are available for each image. In contrast to the FunnyBirds framework, our framework doesn't require training/fine-tuning of the model on the dataset and can be used readily on any model and any method. This framework helps us understand how explanation methods work on part-level. To the best of our knowledge, InG is the first semi-synthetic framework for the evaluation of ViT explanation methods.

Our contributions are twofold: 
\begin{enumerate}[label=\alph*.]
    \item We propose a novel framework called Inpainting the Gaps (InG) to evaluate explanation methods by reducing test-time out-of-distribution-ness, and
    \item We evaluate seven popular ViT explanation methods with our framework to gain an understanding of the usefulness of each method. 
\end{enumerate}

\section{Related Work}
\label{related_work}

\paragraph{General Evaluation Frameworks/Benchmarks}$\mathcal{M}^4$\cite{li2023mathcalm} is a modality-agnostic evaluation benchmark for explanation methods. Adebayo et al. \cite{adebayo2018sanity} proposed data randomization and model parameter randomization tests and found that some explanation methods reconstruct the input. Multiple human-centric evaluation methods \cite{colin2022cannot,kim2022hive} have been proposed as well. \cite{lin2021you} proposed evaluation metrics centred around trojan detection. FunnyBirds \cite{hesse2023funnybirds} uses a synthetic dataset to evaluate the explainability methods. Toolkits like Quantus \cite{hedstrom2023quantus} and OpenXAI \cite{agarwal2022openxai} have been published for evaluating explanation methods. Similar benchmarks were proposed for GNNs \cite{rathee2022bagel} and NLP \cite{wang-etal-2022-fine} as well. Survey papers \cite{nauta2023anecdotal,rong2023towards,shen2020useful} have also been published for the evaluation. 

\paragraph{Methods handling distribution shift}\cite{petsiuk2018rise} uses low-resolution binary maps and upsamples them with bilinear interpolation. \cite{fong2017interpretable} uses blurring on the masked region. \cite{dabkowski2017real} uses U-Net style architecture for generating masks. ROAR\cite{hooker2019benchmark} retrains the network on the perturbed inputs where perturbed pixels are replaced with the per channel mean. However, in ROAR, the model that should be evaluated and the model they are evaluating are different due to retraining. \cite{hsieh2020evaluations} samples values from a uniform distribution for replacing perturbed pixels. \cite{bhalla2023discriminative} uses the normal distribution to inpaint the perturbed pixels. \cite{chang2018explaining} uses Contextual Attention GAN \cite{yu2018generative} to inpaint the mask regions.

\paragraph{Transformers Explainability Methods}Some methods\cite{zhou2016learning,selvaraju2017grad} have been ported directly from CNNs to transformers.  Some differences that might contribute to CNN explainability methods' ineffectiveness on ViTs are self-attention mechanism and GeLUs. Raw Attention Maps(RAM) of the last layer is a simple method for explanation. Attention rollout\cite{abnar2020quantifying} proposed product of attention summed with identity matrix considering skip-attentions as explanation. Attention flow\cite{abnar2020quantifying} considers the network as a graph and calculates the max flow as an explanation. Though proposed for pruning the heads, partial LRP(PLRP)\cite{voita2019analyzing} is an LRP-based method that does LRP till the last layer to find the importance. Transformer Attribution (TA)\cite{chefer2021transformer} and Generic Attribution (GA)\cite{chefer2021generic} explored the usage of relevance propagation in attention mechanism. Other methods like Beyond Intuition (BI)\cite{chen2023beyond} and Transition Attention Maps (TAM)\cite{yuan2021explaining} have proposed theoretical approximations for attention mechanism to explain the model along with integrated gradients\cite{sundararajan2017axiomatic}.

We found \cite{chang2018explaining} to be the nearest neighbor of our work. \cite{chang2018explaining} proposed to inpaint using Contextual Attention GAN\cite{yu2018generative} and then optimize the proposed objective function to generate the saliency map. However, our method, along with inpainting, uses part masks to evaluate explainability rather than generate an explanation map.

\section{Proposed Approach}
\begin{figure}
    \centering
    \includegraphics[width=\linewidth]{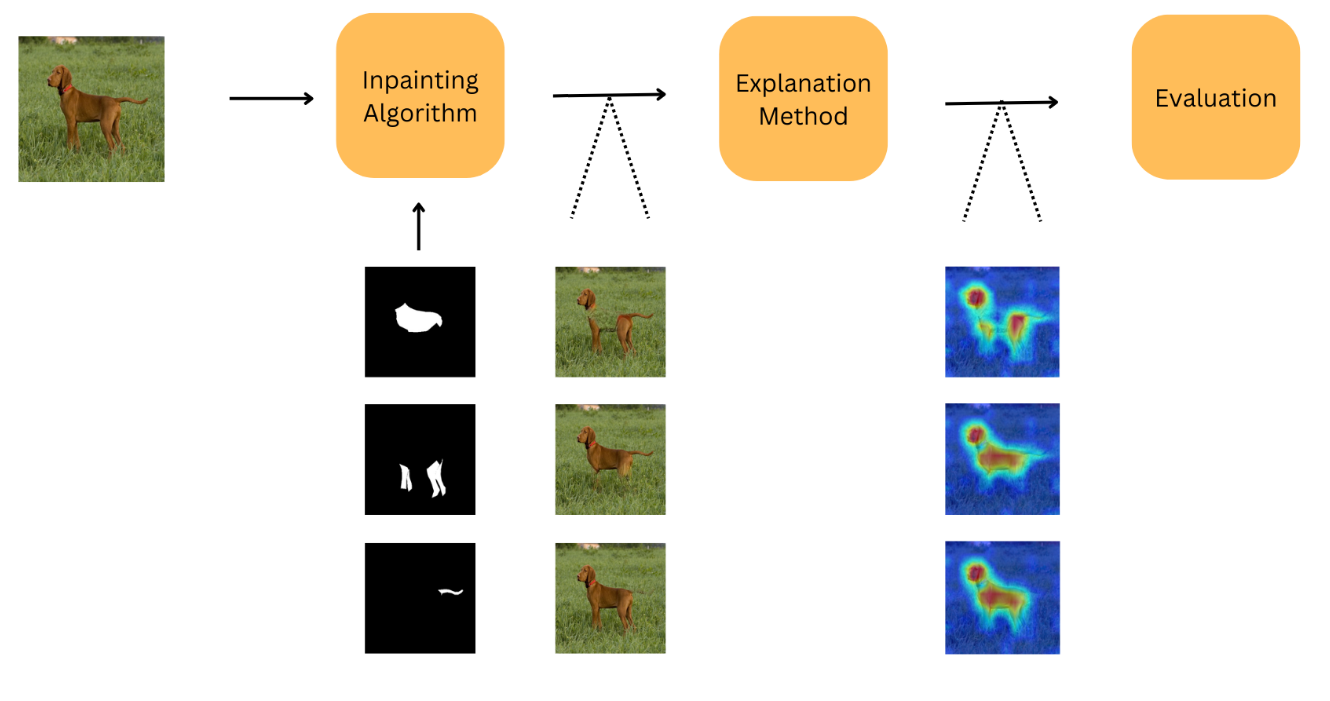}
    \caption{Pipeline of our method. Though the figure shows only single part perturbation, our framework uses multiple-part perturbation. We skipped showing multiple-part perturbation in interest of space.}
    \label{fig:pipeline}
\end{figure}

The FunnyBirds framework \cite{hesse2023funnybirds} has complete control over the dataset generation that allows it to compute the evaluation metrics part-wise. However, this evaluation method is not appropriate with real-world data since the addition/removal of parts would be out-of-distribution. To overcome this problem, we use inpainting models to inpaint the masked regions of the image, thereby reducing the out-of-distribution-ness. Our approach is called Inpainting the Gaps (InG). To understand its advantage, we calculate OTDD~\cite{alvarez2020geometric} between the original dataset (PartImageNet), the dataset generated from pixel masking, and the dataset generated from the inpainting of parts. From Table~\ref{tab:otdd}, it is evident that inpainted images have lesser distributions shift than the images with pixel masking. Figure~\ref{fig:pipeline} shows the pipeline of our method.

We use a part-based dataset, PartImageNet~\cite{he2022partimagenet} which contains annotations of parts. We use MI-GAN~\cite{Sargsyan_2023_ICCV} as our inpainting model to inpaint the parts. This method of evaluating explanations at a concept level rather than pixel level is also connected with the downstream task of providing human-understandable explanations (since humans perceive images in concepts rather than pixels~\cite{hesse2023funnybirds}).

\begin{table}[]
    \centering
    \caption{Dataset distances computed on the PartImageNet dataset}
    \begin{tabular}{ccc}
    \toprule
    Reference dataset & Dataset & OTDD \\
    \midrule
    \multirow{2}{*}{Original Data}
         & Pixel Masking & 85459.00\\
         & Inpainting & \textbf{49026.34}\\
         \bottomrule
    \end{tabular}
    
    \label{tab:otdd}
\end{table}

\subsection{Metrics}\label{metric_section}

\paragraph{Preservation Check (PC):} We pass an image through the model and explanation methods to obtain the predicted class and its explanation. Then, we estimate the importance of each part as the sum of the importance of each pixel in that part. Depending on the importance of each part and the percentage of parts we want to remove, we decide on the order of part removal.

\begin{equation}
    \text{PC}^t = \frac{1}{N}\sum_{n=1}^N \mathbbm{1}[f(x_n^t) = f(x_n)]
\end{equation}
where $\mathbbm{1}[.]$ is the identity function, $x_n^t$ denotes the input $x_n$ with at least $t\%$ importance inpainted where preference is given to least important parts. 

\paragraph{Single Deletion (SD):} Similar to the preservation check, we estimate each part's importance and find the correlation between estimated part importance and actual part importance. Unlike the preservation check, here, we remove only one part at a time, which allows us to find the correlation. This is not the case in multiple-part perturbation as it would be computationally expensive to marginalize for each part.
\begin{equation}
    \text{SD} = \frac12 + \frac{1}{2N} \sum_{n=1}^N \rho(\text{PI}(e_f(x_n)),f(x_n) - f(x_n^t)),
\end{equation}
where $\rho$ is the Spearman rank correlation and part importance (PI) is the sum of importance of pixels of the perturbed part.
\paragraph{Deletion Check (DC):} Deletion check is the inverse of preservation check i.e., instead of removing the unimportant parts, we remove important parts and expect the prediction to change.
\begin{equation}
    \text{DC} = \frac1N \sum_{n=1}^N \mathbbm{1}[f(x_n^t) \neq f(x_n)]
\end{equation}
where $x_n^t$ denotes the input $x_n$ with atleast $t\%$ importance inpainted where preference is given to most important parts.
\paragraph{Perturbation Test:} Similar to \cite{chen2023beyond}, we also perform perturbation test. In contrast to \cite{chen2023beyond}, we perform this on inpainted images and not on pixel-masked images. There are two types of perturbation tests: positive and negative. In the positive perturbation test, we inpaint the parts from high importance to low importance, whereas, in the negative test, the order is reversed. Then, we calculate the accuracy at each level of removal/inpainting and the mean accuracy from them. Since we remove the most important parts in the positive perturbation test, a good explanation method will have a lower score and vice-versa. Again, each test can be conducted for both the predicted class and the target class.

Unlike previous metrics which require a threshold (except SD), the importance of the perturbation test is approximated as the average importance of each pixel in part rather than sum of the importance of each pixel. This is due to the usage of thresholds in previous metrics, which renders per-pixel importance infeasible.

\section{Results and Discussion}
\subsection{Choice of Part Dataset}
There are multiple datasets containing part annotations like PartImageNet \cite{he2022partimagenet}, PACO \cite{ramanathan2023paco}, Pascal-Part \cite{chen2014detect}, ADE20K \cite{zhou2017scene}, Cityscapes-Panoptic Parts\cite{degeus2021panopticparts} etc. PartImageNet is a subset of the ImageNet dataset \cite{5206848}, and contains challenging categories instead of rigid ones. PACO contains more parts than PartImageNet, but some of its images contain more than one object, which makes it unsuitable for classification. We select PartImageNet because it is a subset of ImageNet \cite{5206848} on which the ViT models have been trained. A different dataset choice would require re-training the models on the new dataset. Therefore, to avoid training/fine-tuning the model, we choose PartImageNet.

\subsection{Choice of Inpainting Model}
\begin{figure}
    \centering
    \includegraphics[width=\linewidth]{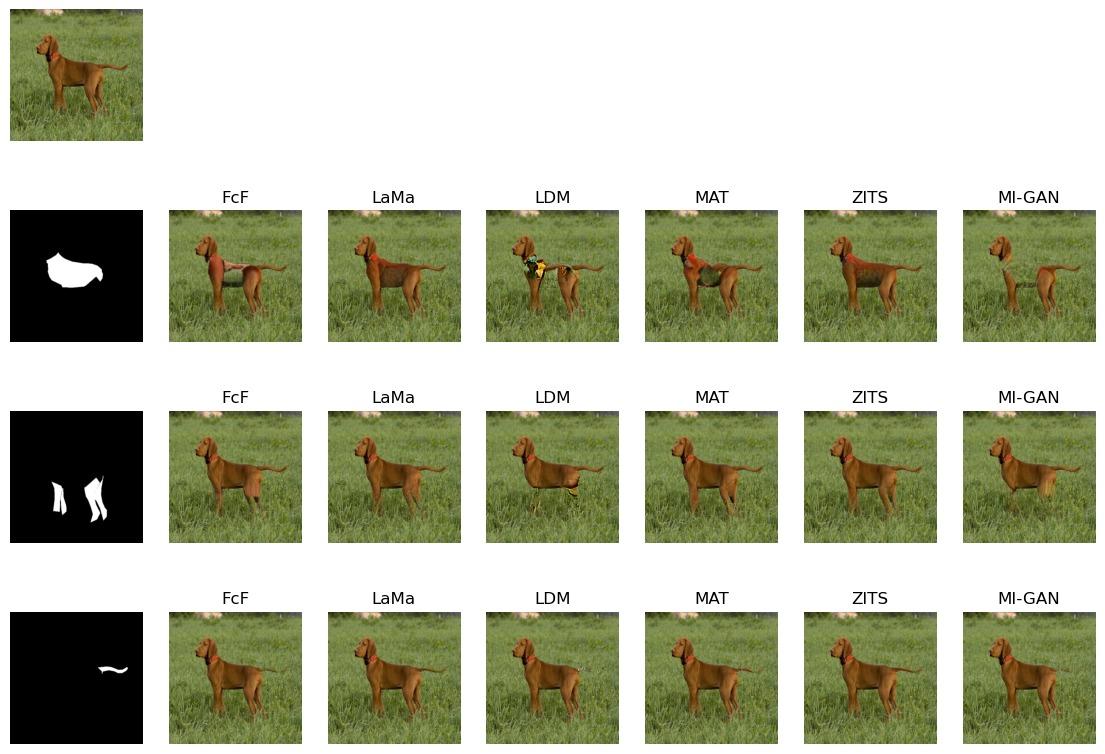}
    \caption{A qualitative example for  of inpainting models. The top-left image is inpainted in the regions identified by the masks in the left column. The label of each inpainted image identifies the inpainting model.}
    \label{fig:inpaint_comparison}
\end{figure}

We qualitatively analyzed FcF \cite{jain2022keys}, LaMa \cite{suvorov2022resolution}, LDM \cite{rombach2022high}, MAT \cite{li2022mat}, MI-GAN \cite{Sargsyan_2023_ICCV} and ZITS \cite{dong2022incremental} for usage in our framework. We found that LaMa and ZITS were able to inpaint the missing parts. However, we want images with those parts removed to be inpainted with the background. We finally chose MI-GAN as it has a lower FID, uses less computation, and generates a plausible image. We use \texttt{iopaint}\footnote{https://github.com/Sanster/IOPaint} to access the inpainting models. Figure \ref{fig:inpaint_comparison} shows a qualitative comparison of the considered inpainting models.

\subsection{Data Generation}
\begin{figure}
    \centering
    \includegraphics[width=\linewidth]{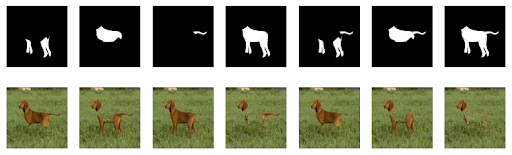}
    \caption{An illustrative example of image generation using the MI-GAN~\cite{Sargsyan_2023_ICCV} model. The top row shows the masks, and the bottom row shows inpainted images. This example shows all possible part removals. Masked regions have blended with the background, resulting in realistic part removal.}
    \label{fig:generation}
\end{figure}

We generate all possible combinations of part masks for a given image. We pass an image and its corresponding parts combination mask to the inpainting model for inpainting. Considering computational resources, we limit the number of combinations per image in PartImageNet to 32. We prefer combinations with fewer parts removal so that at least single deletion protocol has all required images. Figure \ref{fig:generation} presents an illustrative example. 

\subsection{Discussion}
\label{sec:discussion}
Table \ref{tab:perturb_tests} contains results of perturbation tests for some common ViT variants which either change in training strategy or the number of parameters evaluated with our framework. Table \ref{tab:other_metrics} contains results for other metrics defined in Section \ref{metric_section} evaluated with our framework. We use a cluster of 3 NVIDIA RTX A5000 GPUs to run all our experiments. Note that we cannot generate explanations for RAM and Rollout for the target class because they are class-independent.

\begin{table}[]
\centering
\caption{Perturbation Tests (AUC). For positive test, lower is better ($\downarrow$). For negative test, higher is better ($\uparrow$).}
\scalebox{0.87}{
\begin{tabular}{cccccccccccc}
\toprule
     Model & Mode & Class  & RAM          & Rollout          & CAM          & PLRP          & GA           & TA          & BI-Head          & BI-Token \\
     \midrule
     \multirow{4}{*}{\textbf{ViT-Base}} & \multirow{2}{*}{Positive} & Predicted & 23.38 & 22.74 & 24.03 & 23.66 & 23.49 & 23.02 & \textbf{22.29} & 23.09 \\
     & & Target & - & - & 21.35 & 21.53 & 21.02 & 21.08 & \textbf{20.34} & 21.02 \\
& \multirow{2}{*}{Negative} & Predicted & 25.76 & 25.37 & 25.16 & 26.07 & 26.04 & \textbf{27.18} & 26.91 & 26.02 \\
& & Target & - & - & 23.08 & 23.81 & 23.70 & 24.40 & \textbf{24.42} & 23.65 \\
\midrule

\multirow{4}{*}{\textbf{ViT-DINO}} & \multirow{2}{*}{Positive} & Predicted & 21.43 & 22.02 & 21.84 & 21.26 & \textbf{19.92} & 21.60 & 20.54 & 20.68  \\

& & Target & - & - & 20.14 & 19.70 & \textbf{18.50} & 19.78 & 19.06 & 19.22 \\
& \multirow{2}{*}{Negative} & Predicted & 26.91 & 24.62 & 25.95 & 25.92 & \textbf{27.67} & 22.38 & 26.79 & 26.93 \\
& & Target & - & - & 23.66 & 23.74 & \textbf{25.29} & 20.40 & 24.32 & 24.44 \\
     \midrule

\multirow{4}{*}{\textbf{ViT-MAE}} & \multirow{2}{*}{Positive} & Predicted & 28.78 & 28.05 & 30.60 & 25.76 & 26.78 & 26.90 & \textbf{25.60} & 25.72 \\
& & Target & - & - & 29.14 & 24.86 & 25.39 & 25.52 & \textbf{24.74} & 24.87 \\

& \multirow{2}{*}{Negative} & Predicted & 26.68 & 27.83 & 25.88 & 30.72 & 27.07 & 27.01 & 31.62 & \textbf{31.74} \\
& & Target & - & - & 24.91 & 29.35 & 25.88 & 25.85 & 30.19 & \textbf{30.38} \\
\midrule

\multirow{4}{*}{\textbf{ViT-Large}} & \multirow{2}{*}{Positive} & Predicted & 24.14 & \textbf{22.21} & 23.70 & 24.72 & 22.84 & 23.11 & 23.82 & 24.53 \\
& & Target & - & - & 22.53 & 23.78 & \textbf{21.97} & 22.30 & 22.76 & 23.52 \\

& \multirow{2}{*}{Negative} & Predicted & 23.91 & 24.60 & 24.15 & 23.72 & \textbf{26.19} & 25.24 & 24.65 & 24.54 \\
& & Target & - & - & 23.27 & 22.73 & \textbf{25.15} & 24.16 & 23.76 & 23.54 \\
\bottomrule
\end{tabular}
}
\label{tab:perturb_tests}
\end{table}

\begin{table}[]
    \centering
\caption{Comparison of explanation methods. Higher is better ($\uparrow$) for every metric in this table.}   
\begin{tabular}{cccccccccccc}
\toprule
          Model & Metric   & RAM          & Rollout          & CAM          & PLRP          & GA           & TA          & BI-Head          & BI-Token \\ 
\midrule
\multirow{3}{*}{\textbf{ViT-Base}} & SD & 69.45 & 69.76 & 65.24 & 69.74 & \textbf{71.24} & 71.02 & 70.85 & 70.17 \\
& PC & 98.92 & 99.11 & 87.06 & 98.96 & 99.29 & \textbf{99.48} & 99.37 & 99.32 \\
& DC & 78.72 & 82.71 & 50.27 & 81.72 & 82.77 & 82.86 & \textbf{83.23} & 83.03 \\
\midrule
\multirow{3}{*}{\textbf{ViT-DINO}} & SD & 83.52 & 81.66 & 78.14 & 83.08 & \textbf{86.31} & 80.37 & 84.43 & 84.93 \\
& PC & \textbf{99.68} & 99.22 & 93.90 & 99.03 & 99.59 & 98.82 & 99.59 & 99.59 \\
& DC & \textbf{84.15} & 83.23 & 64.76 & 81.91 & 84.10 & 81.31 & 83.81 & 84.06\\
\midrule
\multirow{3}{*}{\textbf{ViT-MAE}} & SD & 74.04 & 74.52 & 66.60 & 74.79 & 74.31 & 74.26 & \textbf{75.72} & 75.91  \\
& PC & 98.59 & 98.93 & 89.74 & 99.02 & 98.61 & 98.56 & \textbf{99.50} & 99.41 \\
& DC & 75.63 & 77.77 & 46.03 & 78.83 & 75.77 & 75.65 & \textbf{80.89} & 80.87 \\
\midrule
\multirow{3}{*}{\textbf{ViT-Large}} & SD & 67.26 & 68.14 & 65.78 & 66.74 & 70.00 & \textbf{68.82} & 68.08 & 67.71  \\
& PC & 98.86 & 99.15 & 49.26 & 98.75 & \textbf{99.53} & 99.41 & 99.30 & 99.28 \\
& DC & \textbf{84.15} & 83.23 & 64.78 & 81.88 & 84.10 & 81.16 & 83.16 & 83.41 \\
\bottomrule
\end{tabular}
    \label{tab:other_metrics}
\end{table}

\begin{figure}[htpb]
    \centering
    \includegraphics[width=0.75\linewidth]{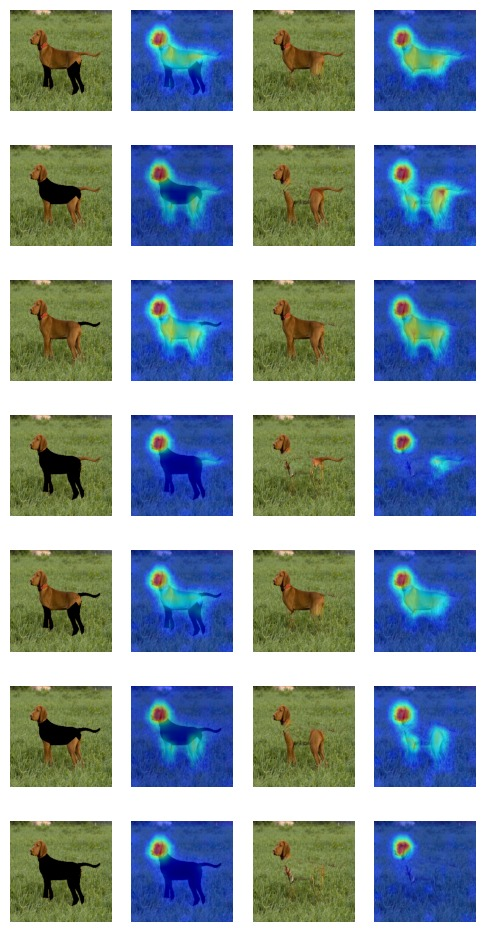}
    \caption{A qualitative comparison of masking and inpainting-based evaluation of the Beyond Intuition-Head (BI) method applied to ViT-Base. The first and third columns are the masked and inpainted images, respectively. The second and fourth columns are the attention maps generated by the BI model. Explanation maps for the masked images show undesired feeble attribution to the masked regions. This is clear from the last row, where the attribution of the masked image includes the masked regions. However, this attribution is much lower in the inpainted images.}    
    \label{fig:qual_exam}
\end{figure}
\paragraph{InG's Impact on Models:} ViT-Base is the most benefitted model from inpainting since it gave a better logit score over pixel masking. This was not the case with DINO and MAE, as there is not much difference in logit scores between masking and inpainting. This highlights the similarity with their training procedures, which use local views and patches instead of whole images, thereby reducing the dependence on the entire object. Observing the explanation of each model, it is clear that every model does not focus on the same part. This behavior is also evident from the different range of scores for each model in Tables \ref{tab:perturb_tests}, \ref{tab:other_metrics}.

\paragraph{Salient Features of InG:} A comparison of masking and InG evaluation of the Beyond Intuition~\cite{chen2023beyond} method is given in Figure~\ref{fig:qual_exam}. It is worth noting that masking retains the shape of the object while removing colour and texture information of the masked part. On the other hand, the inpainted images actually remove these parts and blend them into the background, thereby providing a truer part-based evaluation setting for explanation methods. Further, from Figure \ref{fig:qual_exam}, we see that inpainting results in better attribution in the explanation maps when compared to masking. Table \ref{tab:otdd} provides quantitative evidence for the distribution of inpainted images being closer to natural images when compared to masked images. Though Beyond Intuition delivers superior performance when evaluated through masking, that was not the case with inpainting-based evaluation. This finding shows that some explanation methods gained an advantage with masking-based evaluation.  

\paragraph{Observation on scores:} One interesting point is that raw attention map (RAM) performs better than other complex methods in PC and DC metrics for the DINO model. Even though DINO has segmentation capabilities, that was not the case with pixel masking-based evaluation, as it gave an unfair advantage to other explanation methods. Further, no explanation method can explain every model perfectly according to every performance metric. As in ROAR \cite{hooker2019benchmark}, which re-trains the model to reduce the dataset shift, even in InG, explanation methods perform in a narrow range.

\paragraph{Limitation:} Our work is currently limited to classification setting where images contain single objects and their constituent parts. 

\section{Conclusion}
We find that pixel masking-based evaluation introduces a distribution shift at test time. The test-time distribution shift brings into question the fairness of the pixel-masking evaluation methodology. We proposed Inpainting the Gaps (InG), a framework to overcome this issue. InG is simple, effective and inpaints the masked regions to reduce distribution shift in the test image distribution. Importantly, our method works with real images, unlike recent works that operate on synthetic images. We demonstrated that the InG is more realistic when compared to vanilla pixel-masking both qualitatively and quantitatively (using OTDD). Further, InG can be applied to any ViT explanation method. Through our framework, we uncover some interesting observations on different training strategies. Specifically, we found that GA and BI are competitive explanation methods. Additionally, InG leads to consistently high metrics across various flavours of the ViT. To the best of our knowledge, InG is the first semi-synthetic framework for the evaluation of ViT explanation methods.

\bibliography{refs}

\begin{thebibliography}{10}

\bibitem{abnar2020quantifying}
Samira Abnar and Willem Zuidema.
\newblock Quantifying attention flow in transformers.
\newblock {\em arXiv preprint arXiv:2005.00928}, 2020.

\bibitem{adebayo2018sanity}
Julius Adebayo, Justin Gilmer, Michael Muelly, Ian Goodfellow, Moritz Hardt, and Been Kim.
\newblock Sanity checks for saliency maps.
\newblock {\em Advances in neural information processing systems}, 31, 2018.

\bibitem{agarwal2022openxai}
Chirag Agarwal, Satyapriya Krishna, Eshika Saxena, Martin Pawelczyk, Nari Johnson, Isha Puri, Marinka Zitnik, and Himabindu Lakkaraju.
\newblock Open{XAI}: Towards a transparent evaluation of model explanations.
\newblock In {\em Thirty-sixth Conference on Neural Information Processing Systems Datasets and Benchmarks Track}, 2022.

\bibitem{alvarez2020geometric}
David Alvarez-Melis and Nicolo Fusi.
\newblock Geometric dataset distances via optimal transport.
\newblock {\em Advances in Neural Information Processing Systems}, 33:21428--21439, 2020.

\bibitem{bhalla2023discriminative}
Usha Bhalla, Suraj Srinivas, and Himabindu Lakkaraju.
\newblock Discriminative feature attributions: Bridging post hoc explainability and inherent interpretability.
\newblock In {\em Thirty-seventh Conference on Neural Information Processing Systems}, 2023.

\bibitem{chang2018explaining}
Chun-Hao Chang, Elliot Creager, Anna Goldenberg, and David Duvenaud.
\newblock Explaining image classifiers by counterfactual generation.
\newblock {\em arXiv preprint arXiv:1807.08024}, 2018.

\bibitem{chefer2021generic}
Hila Chefer, Shir Gur, and Lior Wolf.
\newblock Generic attention-model explainability for interpreting bi-modal and encoder-decoder transformers.
\newblock In {\em Proceedings of the IEEE/CVF International Conference on Computer Vision}, pages 397--406, 2021.

\bibitem{chefer2021transformer}
Hila Chefer, Shir Gur, and Lior Wolf.
\newblock Transformer interpretability beyond attention visualization.
\newblock In {\em Proceedings of the IEEE/CVF conference on computer vision and pattern recognition}, pages 782--791, 2021.

\bibitem{chen2023beyond}
Jiamin Chen, Xuhong Li, Lei Yu, Dejing Dou, and Haoyi Xiong.
\newblock Beyond intuition: Rethinking token attributions inside transformers.
\newblock {\em Transactions on Machine Learning Research}, 2023.

\bibitem{chen2014detect}
Xianjie Chen, Roozbeh Mottaghi, Xiaobai Liu, Sanja Fidler, Raquel Urtasun, and Alan Yuille.
\newblock Detect what you can: Detecting and representing objects using holistic models and body parts.
\newblock In {\em Proceedings of the IEEE conference on computer vision and pattern recognition}, pages 1971--1978, 2014.

\bibitem{colin2022cannot}
Julien Colin, Thomas Fel, R{\'e}mi Cad{\`e}ne, and Thomas Serre.
\newblock What i cannot predict, i do not understand: A human-centered evaluation framework for explainability methods.
\newblock {\em Advances in neural information processing systems}, 35:2832--2845, 2022.

\bibitem{dabkowski2017real}
Piotr Dabkowski and Yarin Gal.
\newblock Real time image saliency for black box classifiers.
\newblock {\em Advances in neural information processing systems}, 30, 2017.

\bibitem{degeus2021panopticparts}
Daan de~Geus, Panagiotis Meletis, Chenyang Lu, Xiaoxiao Wen, and Gijs Dubbelman.
\newblock Part-aware panoptic segmentation.
\newblock In {\em IEEE/CVF Conference on Computer Vision and Pattern Recognition (CVPR)}, 2021.

\bibitem{5206848}
Jia Deng, Wei Dong, Richard Socher, Li-Jia Li, Kai Li, and Li~Fei-Fei.
\newblock Imagenet: A large-scale hierarchical image database.
\newblock In {\em 2009 IEEE Conference on Computer Vision and Pattern Recognition}, pages 248--255, 2009.

\bibitem{dong2022incremental}
Qiaole Dong, Chenjie Cao, and Yanwei Fu.
\newblock Incremental transformer structure enhanced image inpainting with masking positional encoding.
\newblock In {\em Proceedings of the IEEE/CVF Conference on Computer Vision and Pattern Recognition}, pages 11358--11368, 2022.

\bibitem{dosovitskiy2020image}
Alexey Dosovitskiy, Lucas Beyer, Alexander Kolesnikov, Dirk Weissenborn, Xiaohua Zhai, Thomas Unterthiner, Mostafa Dehghani, Matthias Minderer, Georg Heigold, Sylvain Gelly, et~al.
\newblock An image is worth 16x16 words: Transformers for image recognition at scale.
\newblock In {\em International Conference on Learning Representations}, 2020.

\bibitem{fong2017interpretable}
Ruth~C Fong and Andrea Vedaldi.
\newblock Interpretable explanations of black boxes by meaningful perturbation.
\newblock In {\em Proceedings of the IEEE international conference on computer vision}, pages 3429--3437, 2017.

\bibitem{he2022partimagenet}
Ju~He, Shuo Yang, Shaokang Yang, Adam Kortylewski, Xiaoding Yuan, Jie-Neng Chen, Shuai Liu, Cheng Yang, Qihang Yu, and Alan Yuille.
\newblock Partimagenet: A large, high-quality dataset of parts.
\newblock In {\em European Conference on Computer Vision}, pages 128--145. Springer, 2022.

\bibitem{hedstrom2023quantus}
Anna Hedstr{\"{o}}m, Leander Weber, Daniel Krakowczyk, Dilyara Bareeva, Franz Motzkus, Wojciech Samek, Sebastian Lapuschkin, and Marina Marina~M.{-}C. H{\"{o}}hne.
\newblock Quantus: An explainable ai toolkit for responsible evaluation of neural network explanations and beyond.
\newblock {\em Journal of Machine Learning Research}, 24(34):1--11, 2023.

\bibitem{hesse2023funnybirds}
Robin Hesse, Simone Schaub-Meyer, and Stefan Roth.
\newblock Funnybirds: A synthetic vision dataset for a part-based analysis of explainable ai methods.
\newblock In {\em Proceedings of the IEEE/CVF International Conference on Computer Vision}, pages 3981--3991, 2023.

\bibitem{hooker2019benchmark}
Sara Hooker, Dumitru Erhan, Pieter-Jan Kindermans, and Been Kim.
\newblock A benchmark for interpretability methods in deep neural networks.
\newblock {\em Advances in neural information processing systems}, 32, 2019.

\bibitem{hsieh2020evaluations}
Cheng-Yu Hsieh, Chih-Kuan Yeh, Xuanqing Liu, Pradeep Ravikumar, Seungyeon Kim, Sanjiv Kumar, and Cho-Jui Hsieh.
\newblock Evaluations and methods for explanation through robustness analysis.
\newblock {\em arXiv preprint arXiv:2006.00442}, 2020.

\bibitem{jain2022keys}
Jitesh Jain, Yuqian Zhou, Ning Yu, and Humphrey Shi.
\newblock Keys to better image inpainting: Structure and texture go hand in hand.
\newblock In {\em WACV}, 2023.

\bibitem{kashefi2023explainability}
Rojina Kashefi, Leili Barekatain, Mohammad Sabokrou, and Fatemeh Aghaeipoor.
\newblock Explainability of vision transformers: A comprehensive review and new perspectives.
\newblock {\em arXiv preprint arXiv:2311.06786}, 2023.

\bibitem{kim2022hive}
Sunnie~SY Kim, Nicole Meister, Vikram~V Ramaswamy, Ruth Fong, and Olga Russakovsky.
\newblock Hive: Evaluating the human interpretability of visual explanations.
\newblock In {\em European Conference on Computer Vision}, pages 280--298. Springer, 2022.

\bibitem{li2022mat}
Wenbo Li, Zhe Lin, Kun Zhou, Lu~Qi, Yi~Wang, and Jiaya Jia.
\newblock Mat: Mask-aware transformer for large hole image inpainting.
\newblock In {\em Proceedings of the IEEE/CVF conference on computer vision and pattern recognition}, pages 10758--10768, 2022.

\bibitem{li2023mathcalm}
Xuhong Li, Mengnan Du, Jiamin Chen, Yekun Chai, Himabindu Lakkaraju, and Haoyi Xiong.
\newblock \${\textbackslash}mathcal\{M\}{\textasciicircum}4\$: A unified {XAI} benchmark for faithfulness evaluation of feature attribution methods across metrics, modalities and models.
\newblock In {\em Thirty-seventh Conference on Neural Information Processing Systems Datasets and Benchmarks Track}, 2023.

\bibitem{lin2021you}
Yi-Shan Lin, Wen-Chuan Lee, and Z~Berkay Celik.
\newblock What do you see? evaluation of explainable artificial intelligence (xai) interpretability through neural backdoors.
\newblock In {\em Proceedings of the 27th ACM SIGKDD conference on knowledge discovery \& data mining}, pages 1027--1035, 2021.

\bibitem{nauta2023anecdotal}
Meike Nauta, Jan Trienes, Shreyasi Pathak, Elisa Nguyen, Michelle Peters, Yasmin Schmitt, J{\"o}rg Schl{\"o}tterer, Maurice van Keulen, and Christin Seifert.
\newblock From anecdotal evidence to quantitative evaluation methods: A systematic review on evaluating explainable ai.
\newblock {\em ACM Computing Surveys}, 55(13s):1--42, 2023.

\bibitem{peebles2023scalable}
William Peebles and Saining Xie.
\newblock Scalable diffusion models with transformers.
\newblock In {\em Proceedings of the IEEE/CVF International Conference on Computer Vision}, pages 4195--4205, 2023.

\bibitem{petsiuk2018rise}
Vitali Petsiuk, Abir Das, and Kate Saenko.
\newblock Rise: Randomized input sampling for explanation of black-box models.
\newblock {\em arXiv preprint arXiv:1806.07421}, 2018.

\bibitem{radford2021learning}
Alec Radford, Jong~Wook Kim, Chris Hallacy, Aditya Ramesh, Gabriel Goh, Sandhini Agarwal, Girish Sastry, Amanda Askell, Pamela Mishkin, Jack Clark, et~al.
\newblock Learning transferable visual models from natural language supervision.
\newblock In {\em International conference on machine learning}, pages 8748--8763. PMLR, 2021.

\bibitem{ramanathan2023paco}
Vignesh Ramanathan, Anmol Kalia, Vladan Petrovic, Yi~Wen, Baixue Zheng, Baishan Guo, Rui Wang, Aaron Marquez, Rama Kovvuri, Abhishek Kadian, et~al.
\newblock Paco: Parts and attributes of common objects.
\newblock In {\em Proceedings of the IEEE/CVF Conference on Computer Vision and Pattern Recognition}, pages 7141--7151, 2023.

\bibitem{rathee2022bagel}
Mandeep Rathee, Thorben Funke, Avishek Anand, and Megha Khosla.
\newblock Bagel: A benchmark for assessing graph neural network explanations.
\newblock {\em arXiv preprint arXiv:2206.13983}, 2022.

\bibitem{rombach2022high}
Robin Rombach, Andreas Blattmann, Dominik Lorenz, Patrick Esser, and Bj{\"o}rn Ommer.
\newblock High-resolution image synthesis with latent diffusion models.
\newblock In {\em Proceedings of the IEEE/CVF conference on computer vision and pattern recognition}, pages 10684--10695, 2022.

\bibitem{rong2023towards}
Yao Rong, Tobias Leemann, Thai-Trang Nguyen, Lisa Fiedler, Peizhu Qian, Vaibhav Unhelkar, Tina Seidel, Gjergji Kasneci, and Enkelejda Kasneci.
\newblock Towards human-centered explainable ai: A survey of user studies for model explanations.
\newblock {\em IEEE Transactions on Pattern Analysis and Machine Intelligence}, 2023.

\bibitem{Sargsyan_2023_ICCV}
Andranik Sargsyan, Shant Navasardyan, Xingqian Xu, and Humphrey Shi.
\newblock Mi-gan: A simple baseline for image inpainting on mobile devices.
\newblock In {\em Proceedings of the IEEE/CVF International Conference on Computer Vision (ICCV)}, pages 7335--7345, October 2023.

\bibitem{selvaraju2017grad}
Ramprasaath~R Selvaraju, Michael Cogswell, Abhishek Das, Ramakrishna Vedantam, Devi Parikh, and Dhruv Batra.
\newblock Grad-cam: Visual explanations from deep networks via gradient-based localization.
\newblock In {\em Proceedings of the IEEE international conference on computer vision}, pages 618--626, 2017.

\bibitem{shen2020useful}
Hua Shen and Ting-Hao Huang.
\newblock How useful are the machine-generated interpretations to general users? a human evaluation on guessing the incorrectly predicted labels.
\newblock In {\em Proceedings of the AAAI Conference on Human Computation and Crowdsourcing}, volume~8, pages 168--172, 2020.

\bibitem{sundararajan2017axiomatic}
Mukund Sundararajan, Ankur Taly, and Qiqi Yan.
\newblock Axiomatic attribution for deep networks.
\newblock In {\em International conference on machine learning}, pages 3319--3328. PMLR, 2017.

\bibitem{suvorov2022resolution}
Roman Suvorov, Elizaveta Logacheva, Anton Mashikhin, Anastasia Remizova, Arsenii Ashukha, Aleksei Silvestrov, Naejin Kong, Harshith Goka, Kiwoong Park, and Victor Lempitsky.
\newblock Resolution-robust large mask inpainting with fourier convolutions.
\newblock In {\em Proceedings of the IEEE/CVF winter conference on applications of computer vision}, pages 2149--2159, 2022.

\bibitem{vaswani2017attention}
Ashish Vaswani, Noam Shazeer, Niki Parmar, Jakob Uszkoreit, Llion Jones, Aidan~N Gomez, {\L}ukasz Kaiser, and Illia Polosukhin.
\newblock Attention is all you need.
\newblock {\em Advances in neural information processing systems}, 30, 2017.

\bibitem{voita2019analyzing}
Elena Voita, David Talbot, Fedor Moiseev, Rico Sennrich, and Ivan Titov.
\newblock Analyzing multi-head self-attention: Specialized heads do the heavy lifting, the rest can be pruned.
\newblock {\em arXiv preprint arXiv:1905.09418}, 2019.

\bibitem{wang-etal-2022-fine}
Lijie Wang, Yaozong Shen, Shuyuan Peng, Shuai Zhang, Xinyan Xiao, Hao Liu, Hongxuan Tang, Ying Chen, Hua Wu, and Haifeng Wang.
\newblock A fine-grained interpretability evaluation benchmark for neural {NLP}.
\newblock In Antske Fokkens and Vivek Srikumar, editors, {\em Proceedings of the 26th Conference on Computational Natural Language Learning (CoNLL)}, pages 70--84, Abu Dhabi, United Arab Emirates (Hybrid), December 2022. Association for Computational Linguistics.

\bibitem{yu2018generative}
Jiahui Yu, Zhe Lin, Jimei Yang, Xiaohui Shen, Xin Lu, and Thomas~S Huang.
\newblock Generative image inpainting with contextual attention.
\newblock In {\em Proceedings of the IEEE conference on computer vision and pattern recognition}, pages 5505--5514, 2018.

\bibitem{yuan2021explaining}
Tingyi Yuan, Xuhong Li, Haoyi Xiong, Hui Cao, and Dejing Dou.
\newblock Explaining information flow inside vision transformers using markov chain.
\newblock In {\em eXplainable AI approaches for debugging and diagnosis.}, 2021.

\bibitem{zhou2016learning}
Bolei Zhou, Aditya Khosla, Agata Lapedriza, Aude Oliva, and Antonio Torralba.
\newblock Learning deep features for discriminative localization.
\newblock In {\em Proceedings of the IEEE conference on computer vision and pattern recognition}, pages 2921--2929, 2016.

\bibitem{zhou2017scene}
Bolei Zhou, Hang Zhao, Xavier Puig, Sanja Fidler, Adela Barriuso, and Antonio Torralba.
\newblock Scene parsing through ade20k dataset.
\newblock In {\em Proceedings of the IEEE Conference on Computer Vision and Pattern Recognition}, 2017.

\end{thebibliography}

\appendix

\section{Supplementary Material}

\begin{figure}[htpb]
    \centering
    \includegraphics[width=\linewidth]{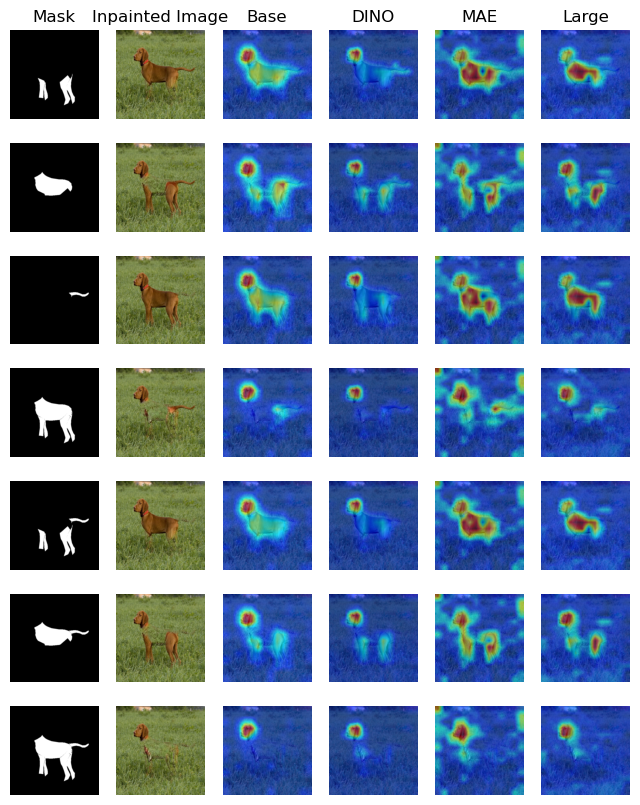}
    \caption{A qualitative example of how explanation methods work on inpainted images. Explanations are generated with Beyond Intuition-Head. The first two columns are mask and inpainted images, respectively. The remaining columns are the explanation maps generated by models, which are mentioned as the title of the column.}    
    \label{fig:inpaint_vit_dog}
\end{figure}

\begin{figure}[htpb]
    \centering
    \includegraphics[width=\linewidth]{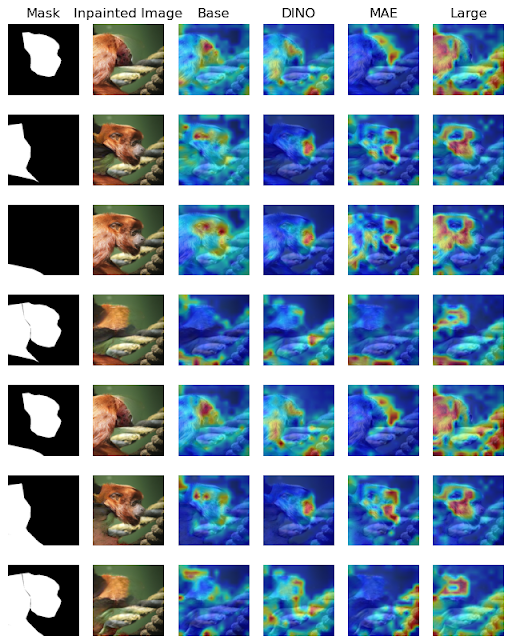}
    \caption{}    
    \label{fig:inpaint_vit_monkey}
\end{figure}

\begin{figure}[htpb]
    \centering
    \includegraphics[width=\linewidth]{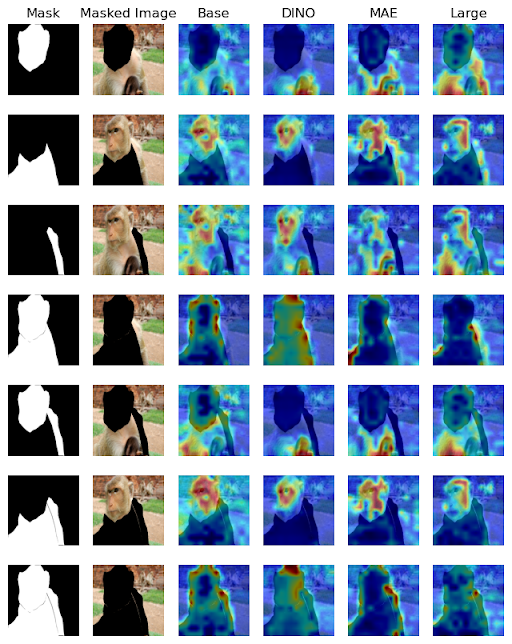}
    \caption{}    
    \label{fig:mask_vit_boat}
\end{figure}

\begin{figure}[htpb]
    \centering
    \includegraphics[width=\linewidth]{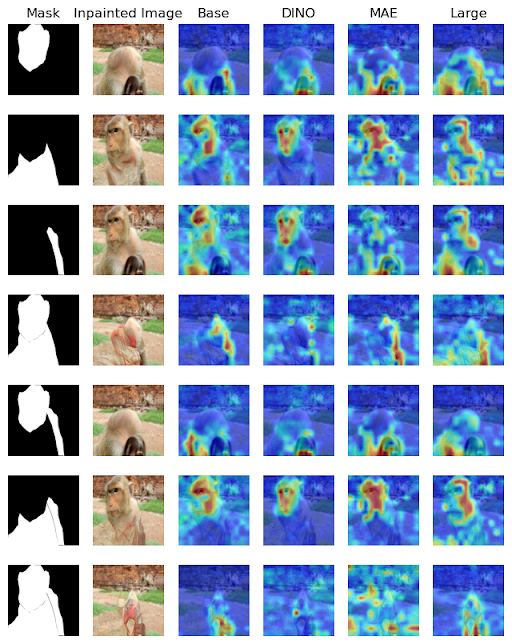}
    \caption{}    
    \label{fig:inpaint_vit_boat}
\end{figure}

Images start from the next page. Images in the supplementary material are generated with respect to the predicted class and not the ground truth class. From Figure \ref{fig:mask_vit_boat}, it is clear that explanation methods focus on unrealistic regions(masked pixels). This highlights the drawback of pixel masking for evaluation. This is not found in the inpainting based evaluation which can be seen in Figure\ref{fig:inpaint_vit_dog}\ref{fig:inpaint_vit_monkey}\ref{fig:inpaint_vit_boat}

\begin{figure}[htpb]
    \centering
    \includegraphics[width=\linewidth]{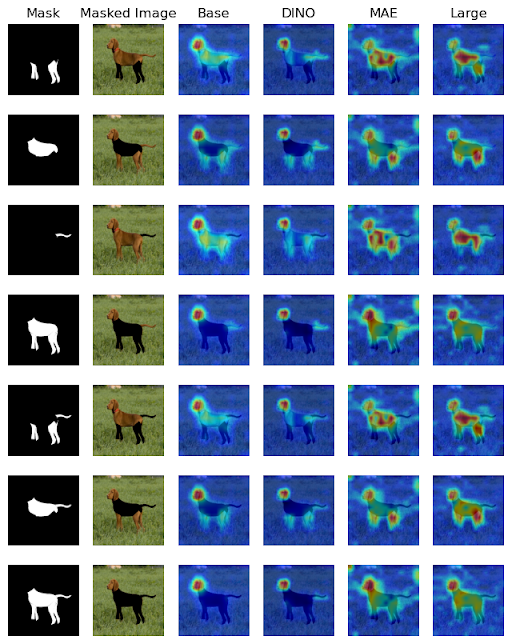}
    \caption{A qualitative example of how explanation methods work on masked images. Explanations are generated with Beyond Intuition-Head. The first two columns are mask and masked images, respectively. The remaining columns are the explanation maps generated by models, which are mentioned as the title of the column.}    
    \label{fig:mask_vit_dog}
\end{figure}
\begin{figure}[htpb]
    \centering
    \includegraphics[width=\linewidth]{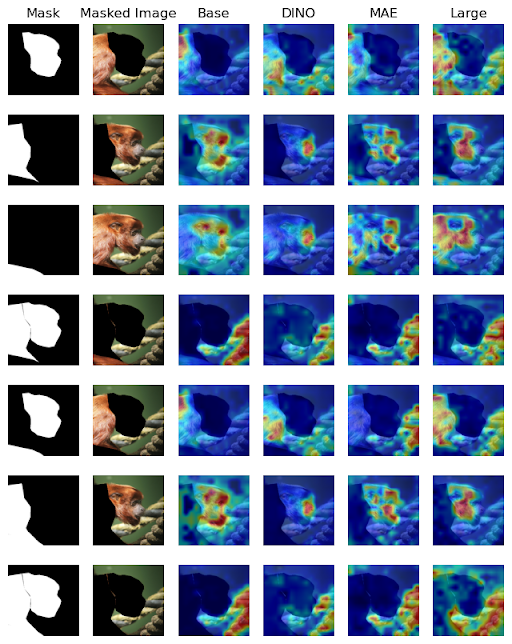}
    \caption{}    
    \label{fig:mask_vit_monkey}
\end{figure}

\end{document}